\documentclass{article}
\usepackage{spconf,amsmath,graphicx}
\usepackage{graphicx}
\usepackage{lipsum}
\usepackage{bm}
\usepackage{gensymb}
\usepackage{amssymb}
\usepackage{pifont}
\newcommand{\cmark}{\ding{51}}%
\newcommand{\xmark}{\ding{55}}%

\usepackage{enumitem}

\title{DUODEPTH: STATIC GESTURE RECOGNITION WITH DUAL DEPTH SENSORS}
\name{Ilya Chugunov and Avideh Zakhor}
\address{University of California, Berkeley}
\usepackage{setspace}

\begin{document}
%
\maketitle
\begin{abstract}
Static gesture recognition is an effective non-verbal communication channel between a user and their devices; however many modern methods are sensitive to the relative pose of the user's hands with respect to the capture device, as parts of the gesture can become occluded. We present two methodologies for gesture recognition via synchronized recording from two depth cameras to alleviate this occlusion problem. One is a more classic approach using iterative closest point registration to accurately fuse point clouds and a single PointNet architecture for classification, and the other is a dual PointNet architecture for classification without registration. On a manually collected data-set of 20,100 point clouds we show a 39.2\% reduction in misclassification for the fused point cloud method, and 53.4\% for the dual PointNet, when compared to a standard single camera pipeline.
\end{abstract}
\begin{keywords}
Gesture recognition, point cloud, light-weight, occlusion
\end{keywords}
\section{Introduction}
\label{sec:intro}
With the ever increasing prevalence of human-computer interaction tasks, it is no surprise that gesture recognition remains a major topic of research in computer vision \cite{review, 2016survey, survey}. With any vision task occlusion means a loss of data, and thus degraded performance. Even without external disturbances, such as a plant in front of the camera \cite{integral-imaging}, hand gestures have a tendency to self-occlude if their pose does not line up well with the geometry of the recording camera, as seen in Fig.1. For robust operation this hole in the data must be patched with another source of information. 

In current literature there are three main approaches to the occlusion problem in gesture recognition: intra-frame analysis, inter-frame analysis, and sensor fusion. The first approach is well demonstrated in \cite{integral-imaging} and \cite{compressive-sensing}, where via processes such as integral imaging and compressed sensing the authors are able to extract more salient information directly from single frames, combating occlusion with algorithms that are naturally robust to it. Inter-frame analysis can be seen in papers such as \cite{temporal} and \cite{spatiotemporal} wherein multiple frames are used to provide better contextual information for gestures, reducing the impact of temporary occlusions. The sensor fusion approach is the conceptually simplest but least explored \cite{sensor-noise}. 

\begin{figure}[htb] 
    \centering
    \includegraphics[width=0.45\textwidth]{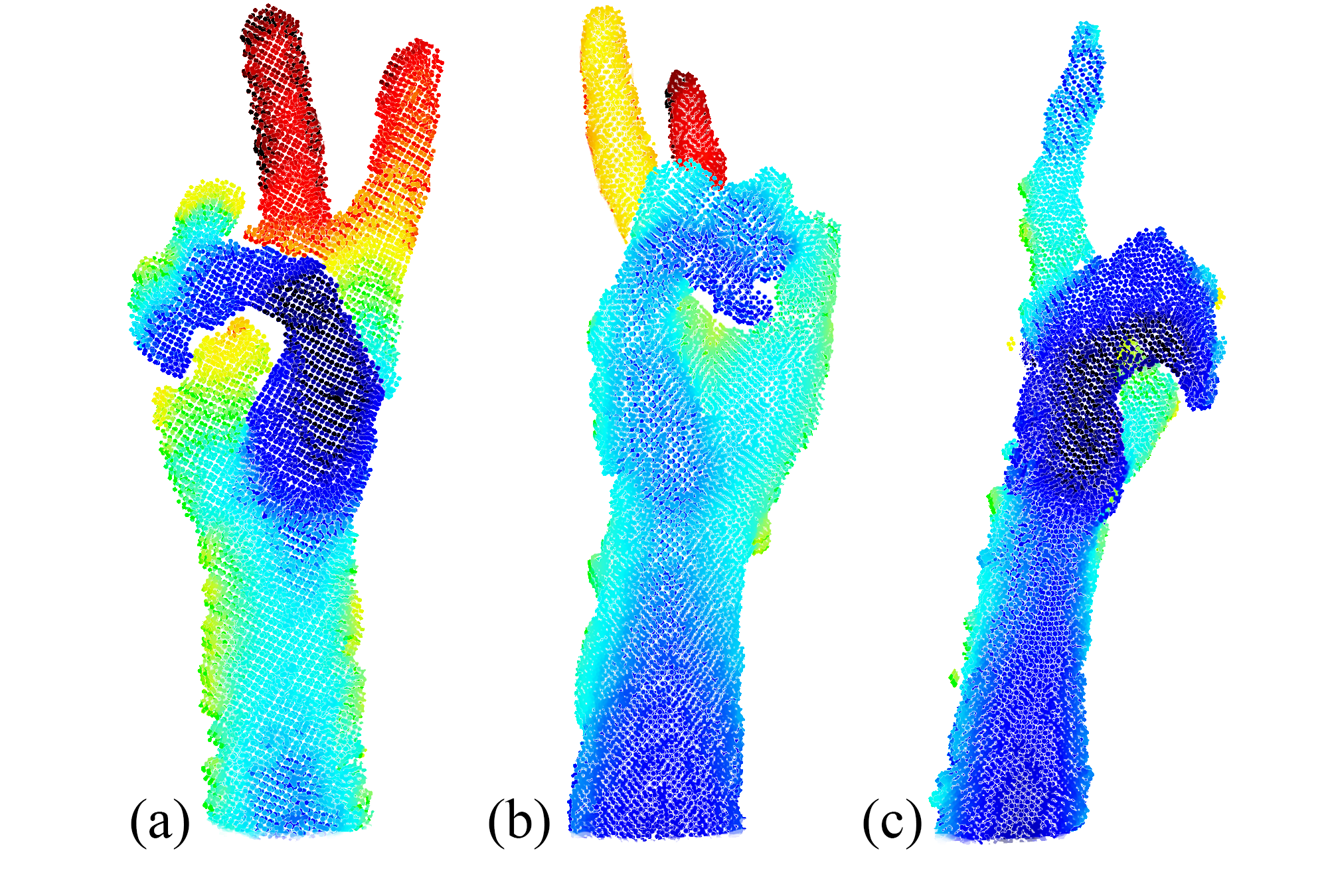}
    \caption{Two gesture facing towards (a) left camera; (b) neither camera; and (c) right camera. Recorded from left camera.}
    \vspace{-1em}
    \label{twofinger}
\end{figure}

\noindent
 For this paper we focus on intra-frame analysis and sensor fusion as there is limited temporal information to be gained from static gestures. Whether with multiple of the same sensor \cite{multisensor}, or with a diverse combination \cite{leap, multisensor2}, the primary problem of sensor fusion remains the same, how to best process the multiple streams of data. We propose using the PointNet architecture \cite{pointnet}, which has been shown to perform well in problems of hand pose estimation \cite{hand-pointnet,noaugmentation1} as well as general object recognition \cite{pointnet,pointnet++,frustum}. We opt to use point clouds for the collected hand data as they are light-weight, allowing for a sparse representation of recorded data without wasting points on empty space \cite{pointnet, frustum}, and provide an intuitive way in which to fuse data from multiple camera pipelines, via coordinate system transformation and concatenation.

In this paper we outline two primary methodologies for this problem, and characterize their performance against single camera solutions: \vspace{-0.5em}
\begin{enumerate}[noitemsep]
    \itemsep 0.1em
    \item Gesture recognition via single PointNet architecture on fused features derived from point clouds fused with ICP registration, referred to as FUSED.
    \item Gesture recognition via dual PointNet architecture on:
    \begin{enumerate}
    \item fused features derived from two separate point clouds, referred to as DUAL-FEAT.
    \item independent features derived from two separate point clouds, referred to as SOLO-FEAT. 
  \end{enumerate}
\end{enumerate} 
\vspace{-1em}


\begin{figure}[htb] 
    \centering
    \includegraphics[width=0.45\textwidth]{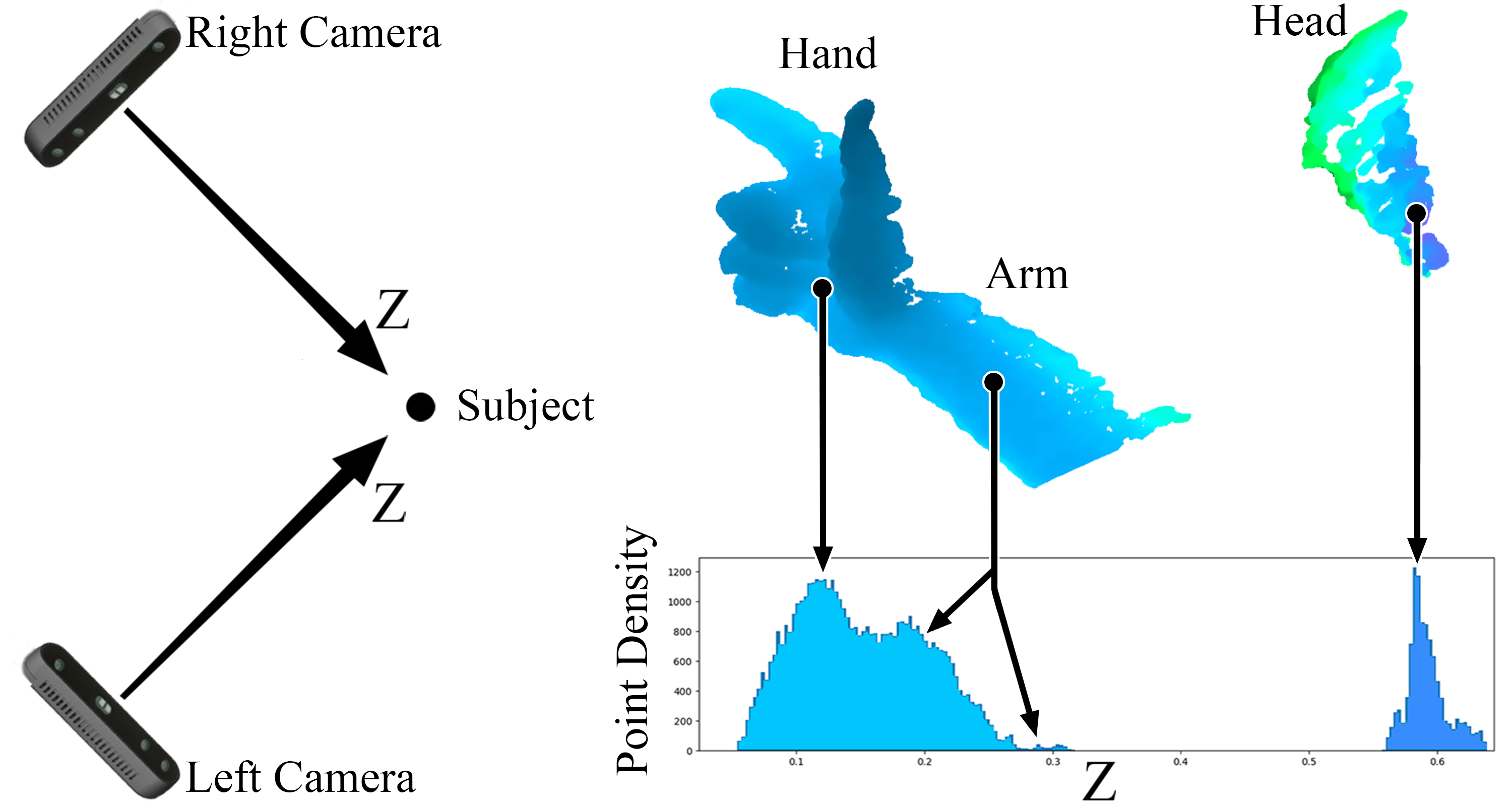}
    \vspace{-5px}
    \caption{Dual Intel RealSense D415 setup (left), depth capture and histogram of point density along Z axis (right).}
    \vspace{-1em}
    \label{recovery}
\end{figure}

\section{DuoDepth}
\vspace{-0.5em}

\subsection{Data collection}
\label{sec:format}
 We manually recorded gestures with two depth cameras mounted perpendicular to each-other and angled 45 $\degree$ upward with respect to the horizontal, as seen in Fig.2. 1005 captures were taken of each of ten gestures, divided into 335 captures of the gesture facing towards the left camera, 335 facing the right camera, and 335 facing towards the laptop screen, i.e. neither camera. Each capture was comprised of two point clouds, one from each camera. An example of these point clouds can be seen in Fig.1. Both depth cameras were depth limited for recording gestures, to minimize background noise, however at the start of each recording session several depth unlimited captures of the surrounding area were taken for registration purposes. The 1005 captures, two point clouds per capture, and ten gestures resulted in a total of 20,100 recorded point clouds which were then divided 80-20\% into the train and test set. Unlike \cite{integrating} we do not use skin color for segmentation, nor does PointNet use color, thus the gestures' RGB data is discarded.
\subsection{Data fusion}
\label{sec:format}
In order to fuse the point clouds in each recording session for the FUSED architecture, we use an initial transform derived from the geometry of the camera setup as the seed for Open3D's color iterative closest point (ICP) function \cite{open3d}. This algorithm, derived from \cite{Park2017ColoredPC}, is run on the left and right depth unlimited captures and seeks to optimize:
\begin{equation}
E(T)=(1-\delta)E_C(T)+\delta E_G(T)
\end{equation}
where T is the transformation matrix, $\delta$ is a weight parameter, $E_C$ is photometric error, and $E_C$ is geometric error as defined in Point-to-Plane ICP \cite{CHEN1992145}. This allows us to find an accurate transform between the coordinate systems of the left and right depth camera for each session, which can then be used to concatenate the point clouds. If only the initial geometric transform is used without ICP refinement, then small movements in camera cables and other disturbances can cause point cloud misalignments millimeters in magnitude.

\begin{figure}[htb] 
    \centering
    \includegraphics[width=0.5\textwidth]{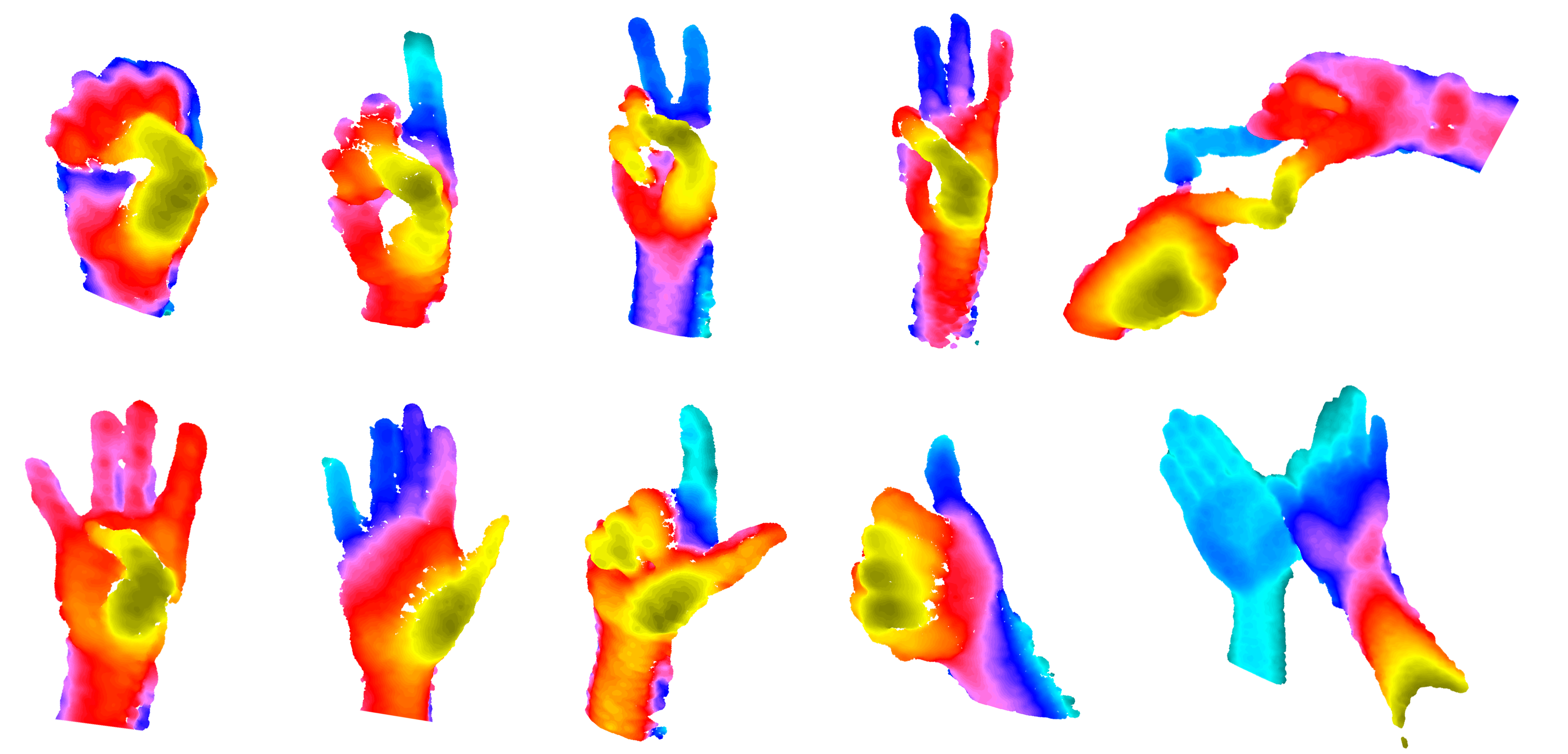}
    \vspace{-15px}
    \caption{The ten recorded gestures. Top left to bottom right: Zero, One, Two, Three, Frame, Four, Five, Ell, Thumb, Bird.}
    \vspace{-1em}
    \label{gestures}
\end{figure}
\vspace{-3px}
\subsection{Point cloud processing}
Although the maximum depth of the recording cameras is limited in the hardware settings, ensuring objects behind and around the user are not captured, the user's face and body are often still present in the final point clouds. Applying a hand isolation technique similar to \cite{arm-removal} would not work until the samples are stripped of these body and face artifacts. We found empirically that the distribution of recorded points along the $Z$ axis provides the necessary information for cropping. Specifically there exists a high density of points along the $Z$ axis at the face of the gestures, such as the palm for the One gesture, as shown in Fig.2. There also appears a high density of points at the location of the body/face, if it was captured, which leads to a multi-modal distribution of points along the $Z$ axis. Thus, applying:
\begin{equation}
    detect\_peaks(data_Z)\rightarrow
    \begin{cases}
        \#peaks=1,& \text{\textbf{pass}}\\
        \#peaks=2+,& \text{\textbf{isolate arm}}\\
    \end{cases}
\end{equation}
would result in point clouds of only the user's arm, which could then undergo fine processing \cite{arm-removal} to isolate the hand. Here $data_Z$ can be any measure of point density along the $Z$ axis, such as a histogram on the $Z$ data of the point cloud, and \textbf{isolate arm} means to discard all contiguous clusters of points beyond the first detected peak.

Similar to \cite{pointnet, augmentation1, augmentation2} we found that data augmentation proved to be beneficial for classification accuracy, helping combat over-fitting in high-epoch training. Points were randomly jittered as in \cite{pointnet}, and random translations were performed on the point cloud as a whole. Rotational augmentation was not used, as even at small magnitudes of $<5\degree$ it invariably lowered test accuracy. Thus the augmentation of a point cloud with points at $(X^N,Y^N,Z^N)$ takes the form:
\vspace{-0.5em}
\begin{equation}
\begin{gathered}
   \widetilde{P}^N= (X^N,Y^N,Z^N) + (J^N_x,J^N_y,J^N_z) + (T_x,T_y,T_z)^N\\
   T_x,T_y,T_z\sim \mathcal{N}(0,\alpha)\hspace{2em}J^N_x,J^N_y,J^N_z \sim \mathcal{N}_N(0, \beta)
\end{gathered}
\end{equation}
where $\widetilde{P}=(\tilde{X}^N,\tilde{Y}^N,\tilde{Z}^N)$ is the new point cloud of size $N$, J is the jitter size, T is the translation size, and $\alpha$ and $\beta$ are empirically determined as 0.002 and 0.01 in Section 3.2.

\begin{figure}[htb] 
    \centering
    \includegraphics[height=0.34\textwidth]{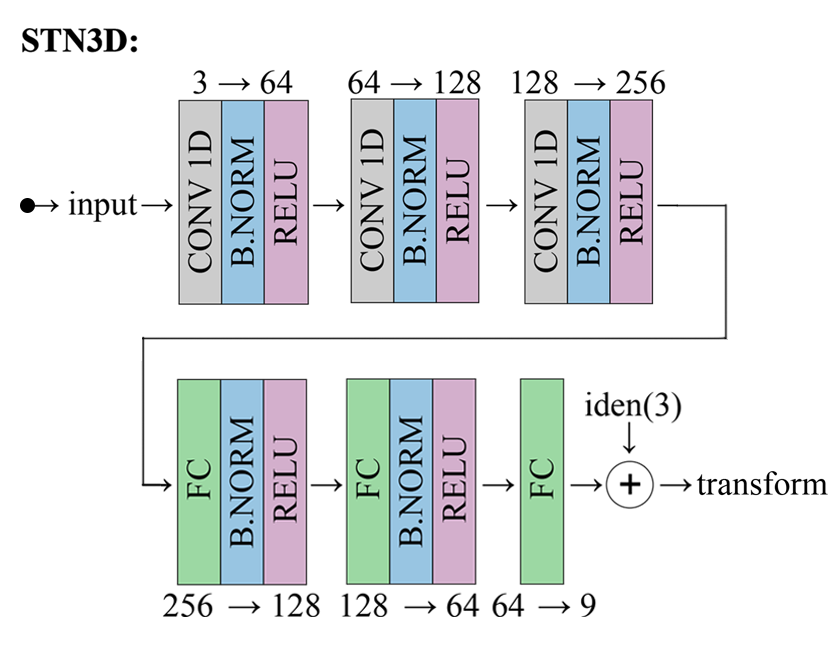}
    \vspace{-15px}
    \caption{3D spatial transformer network (\textbf{STN3D}).}
    \vspace{-1em}
    \label{recovery}
\end{figure}

\subsection{Network architecture}
The network architectures in this paper are each constructed from three base components: the 3D spatial transformer network, feature extractor, and classifier. The transformer network, shown in Fig.4, is a modified version of STN3D \cite{stn3d-git}, which itself is a 3D adaptation of the methods of \cite{stn3d}. It outputs per-sample transformations to be used in the feature extraction layers, and serves to produce a final model invariant to rotations, translations, and changes of scale; acting as a T-net \cite{pointnet}. The feature extractor (FEAT) and classifier layers (CLS), shown in Fig.5, are direct adaptations of the PointNet architecture \cite{pointnet} \cite{pytorch-git}, with significantly reduced layer sizes.

\subsubsection{Fused features, fused inputs (\textbf{FUSED})}
For the FUSED architecture $input$ is a single fused point cloud containing information from both the left and right cameras, as described in Section 2.3. We thus use a single PointNet pipeline:
\begin{equation}
\begin{gathered}
input \searrow \hspace{6.5em} \\
input \rightarrow STN3D \rightarrow FEAT \rightarrow CLS \rightarrow output
\end{gathered} 
\end{equation}
A 3D spatial transform is calculated for the fused point cloud as a whole, and this transform is passed along with the original $input$ into the feature extraction and classification layers.

\subsubsection{Fused features, separate inputs (\textbf{DUAL-FEAT})}
In the DUAL-FEAT architecture $input1$ and $input2$ are the two individual point clouds produced by the two cameras:
\begin{equation}
\begin{gathered}
input1 \searrow \hspace{6.5em} \\
input1 \rightarrow STN3D \rightarrow FEAT \rightarrow CLS \rightarrow output \\
input2 \rightarrow STN3D \nearrow \hspace{11.5em} \\
input2 \nearrow \hspace{6.5em}
\end{gathered} 
\end{equation}
3D spatial transforms are calculated independently for each input point cloud, which are used to fuse the inputs before being passed into the feature extraction and classification layers.

\begin{figure}[htb] 
    \centering
    \includegraphics[height=0.34\textwidth]{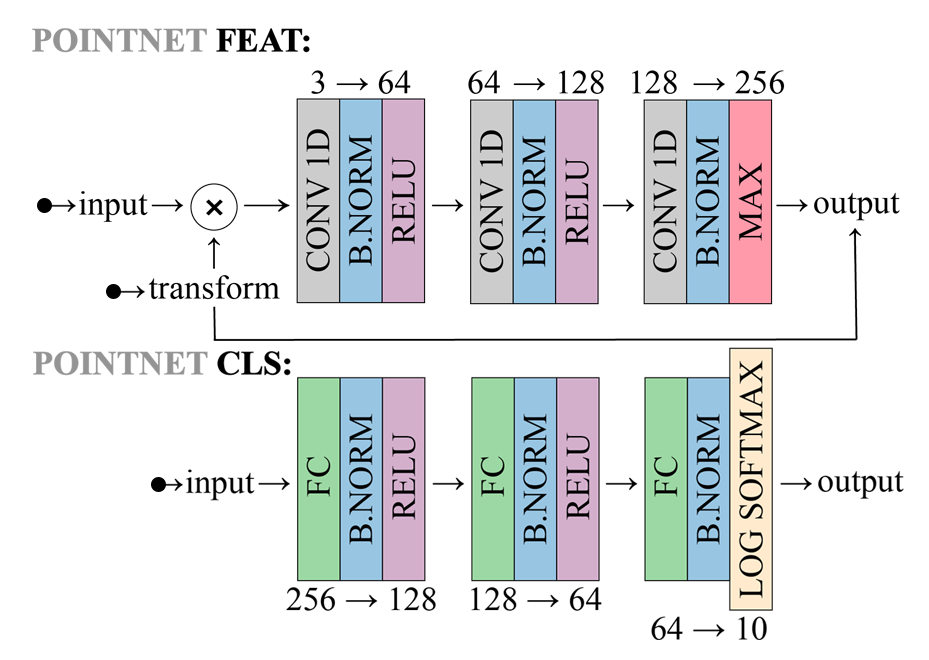}
    \vspace{-15px}
    \caption{Feature extractor (\textbf{FEAT}) and classifier (\textbf{CLS}).}
    \vspace{-1em}
    \label{recovery}
\end{figure}

\subsubsection{Independent features, separate inputs (\textbf{SOLO-FEAT})}
In SOLO-FEAT, $input1$ and $input2$ are again two individual point clouds produced by the two cameras:
\begin{equation}
\begin{gathered}
input1 \searrow \hspace{6.5em} \\
input1 \rightarrow STN3D \rightarrow FEAT \rightarrow CLS \rightarrow output \\
input2 \rightarrow STN3D \rightarrow FEAT \nearrow \hspace{7em} \\
input2 \nearrow \hspace{6.5em}
\end{gathered}
\end{equation}
This architecture is similar to (5), however inputs are never fused and instead undergo separate feature extraction. Their features are then concatenated and passed into the classifier.

\section{EXPERIMENTS}
\vspace{-0.25em}
\subsection{Implementation and runtime}
Network code is adapted from \cite{pytorch-git}, a re-implementation of PointNet in PyTorch, and was executed on a machine with an Intel Core i7-6850K CPU @ 3.60GHz and a single GeForce GTX 1080Ti. Code and data-set are available at \cite{my-git}.

\setcounter{table}{1}
\begin{table*}[htb]
\centering
\resizebox{\textwidth}{!}{\
\begin{tabular}{|c|c|c|c|c|c|c|c|c|c|c|c|}
\hline
Pipeline & Overall & Zero & One & Two & Three & Four & Five & Thumb & Ell & Frame & Bird\\
Variant & Mean & Mean & Mean & Mean & Mean & Mean & Mean & Mean & Mean & Mean & Mean\\
\hline
Left Only& 0.878 & 0.873 & 0.840 & 0.939 & 0.905 & 0.850 & 0.724 & 0.934 & 0.828 & 0.968 & 0.921 \\
\hline
Right Only & 0.885  & 0.886 & 0.854 & 0.840 & 0.880 & 0.867 & 0.764 & 0.954 & 0.889 & 0.976 & 0.937\\
\hline
FUSED & 0.928  & 0.968 & 0.909 & 0.918 & 0.893 & 0.942 & \textbf{0.801} & 0.971 & 0.933 & 0.990 & 0.954 \\
\hline
DUAL-FEAT & \textbf{0.945}  & \textbf{0.972} & \textbf{0.931} & \textbf{0.975} & 0.941 & \textbf{0.979} & 0.771 & 0.970 & \textbf{0.953} & 0.995 & \textbf{0.963} \\
\hline
SOLO-FEAT & 0.938 & 0.938 & 0.919 & 0.971 & \textbf{0.967} & 0.932 & 0.791 & \textbf{0.972} & 0.932 & \textbf{0.997} & 0.962 \\
\hline

\end{tabular}
}
\caption{Accuracy results for various inputs and architectures for: 100 trials, 100 epochs, 8040 train, 2010 test}
\vspace{-1em}
\end{table*}

All pipelines required an average of 2.9ms to crop, 3.0ms to down-sample, and 1.2ms to classify a single input. A total of 7.1ms per input, significantly lower than the 20ms+ required by works such as \cite{projection,albanian}. Cropping and down-sampling times can be further improved to under a millsecond each by limiting point cloud acquisition size in the hardware of the cameras, however the effect of this on overall classification accuracy remains untested.

\subsection{Ablation testing}
In \cite{pointnet}, the authors found that classification accuracy improved as the number of points in the input point clouds increased, however noted little to no improvement from using over one thousand points. We ran a similar series of tests, displayed in Fig.6, and observed improvements in overall classification accuracy up to a point cloud size of 320, which was chosen as $N$, the size for training and testing. Augmen-

\begin{figure}[htb] 
    \centering
    \includegraphics[width=0.47\textwidth]{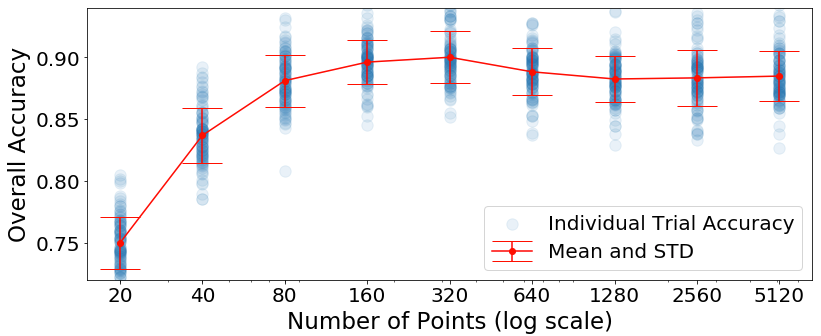}
    \caption{Point cloud size versus mean accuracy for: 100 trials, 32 epochs, 8040 train, 2010 test, and FUSED architecture.}
    \vspace{-1em}
    \label{size}
\end{figure}
\noindent
tation magnitudes $\alpha$ and $\beta$, used in Equation (3), were chosen via a similar process. 

As shown in Table 1 a number of augmentations of the original PointNet architecture were tested, including: (D) dropout in the second fully connected layer in the classifier shown in Fig.5, (R) reduction of convolution and fully connected layer sizes, and (SELU) self normalizing functions in place of batch normalization and RELU \cite{SELU}. Due to our use of smaller point clouds than in \cite{pointnet} for training and testing, the largest improvement in accuracy was achieved from reduction of the layer sizes, or equivalently the number of network parameters.
\subsection{Results}
Table 2 showcases the final test results of the three dual camera pipelines as well as two single camera pipelines, which were both trained on the standard PointNet architecture in Equation (4). The left and right cameras performed similarly overall, however presented noticeable classification disparities in more asymmetric gestures such as Two, Five, and Ell. The FUSED pipeline produced a 39.2\% reduction in misclassification when compared to averaged single camera performance. DUAL-FEAT and SOLO-FEAT showed higher 53.4\% and 47.7\% reductions in misclassification respectively, with DUAL-FEAT achieving top performance for the majority of gestures. This is likely due to its ability to extract combined features from a fused point cloud, correctly classifying captures where both the left and right camera receive occluded data. In these cases SOLO-FEAT would extract incomplete features from each occluded input individually and with them produce an incorrect classification.

\setcounter{table}{0}
\begin{table}[htb]
\centering
\resizebox{\columnwidth}{!}{%
\begin{tabular}{|c|c|c|c|c|c|}
\hline
Augmentation: & NONE & D & R & R+D & SELU\\
\hline
Mean Accuracy:  & 0.899 & 0.903 & \textbf{0.907} & 0.885 & 0.858 \\
\hline
\end{tabular}}
\caption{Network augmentation ablation results for: 300 trials, 64 epochs, 8040 train, 2010 test, and FUSED architecture.}
\vspace{-1em}
\end{table}

\setcounter{table}{2}
\begin{table}[htb]
\resizebox{\columnwidth}{!}{%
\centering
\begin{tabular}{|c|c|c|c|c|c|}
\hline
Method: & \cite{projection} &  DUAL-FEAT & \cite{attention} & \cite{leap} & \cite{spatiotemporal} \\
\hline
Mean Accuracy: & \textbf{0.999} & 0.945 &  0.942 & 0.913 & 0.844 \\
\hline
Train Size: & 28K & 20.1K & 26.7K & 1.4K  & 26.7K \\
\hline
Data-set Available: & \xmark & \cmark & \cmark & \cmark & \cmark \\
\hline
\end{tabular}
}
\caption{Comparison of mean accuracy for 10 static gesture classification.}
\vspace{-1.5em}
\end{table}

\noindent
When compared to other modern methods for static gesture recognition with depth data, as shown in Table 3, our single camera test results are noncompetitive, as the data-set in this paper is purposely created to be challenging for a single view pipeline. The DUAL-FEAT mean accuracy is however quite similar to \cite{attention}, which uses 4 point clouds collected temporally from a single depth camera for classification. The point cloud samples in \cite{attention} are created with the test subject always facing towards the camera, favourably to its geometry, and as such the methods of \cite{attention} are not likely to perform as well on a more challenging self-occluded data-set such as the one presented in this paper.  Additionally DuoDepth uses 12.8 times fewer points per input than \cite{attention}, requiring a lower number of network parameters to train and significantly less computation per classification.

\vspace{-1em}

\begin{table}[htb]
\resizebox{\columnwidth}{!}{%
\centering
\begin{tabular}{|c|c|c|c|c|c|}
\hline
Max Rotation ($\zeta$): & $1\degree$ & $5\degree$ & $10\degree$ & $15\degree$ & $20\degree$\\
\hline
FUSED: & 0.854 & 0.555 & 0.499 & 0.334 & 0.313 \\
\hline
DUAL-FEAT:  & 0.838 & 0.541 & 0.415 & 0.347 & 0.340 \\
\hline
SOLO-FEAT:  & \textbf{0.907 }& \textbf{0.767 }&\textbf{ 0.836} & \textbf{0.838} & \textbf{0.849} \\
\hline

\end{tabular}}
\caption{Rotational augmentation mean accuracy results for: 100 trials, 32 epochs, 8040 train, 2010 test.}
\vspace{-1em}
\end{table}
\noindent
As mentioned in Section 2.3, adding rotational augmentation invariably lowered test accuracy. However by applying random rotations of magnitude $\mathcal{N}(0, \zeta)$ to left and right inputs independently it was observed that SOLO-FEAT had a significantly higher resilience to rotational noise than the other two architectures, as shown in Table 4. This presents interesting potential applications for SOLO-FEAT, such as multi-robot gesture recognition. If there are no good estimates of relative poses between robots' on-board cameras, then random rotations naturally appear in the data, which could lead to degradation in gesture recognition and general performance.

\bibliographystyle{IEEEbib}
\small{
\bibliography{refs}

\begin{thebibliography}{10}

\bibitem{review}
Rafiqul~Zaman Khan and Noor Ibraheem,
\newblock ``Hand gesture recognition: A literature review,''
\newblock {\em International Journal of Artificial Intelligence \& Applications
  (IJAIA)}, vol. 3, pp. 161--174, 08 2012.

\bibitem{2016survey}
Hong Cheng, Lu~Yang, and Zicheng Liu,
\newblock ``Survey on 3d hand gesture recognition.,''
\newblock {\em IEEE Trans. Circuits Syst. Video Techn.}, vol. 26, no. 9, pp.
  1659--1673, 2016.

\bibitem{survey}
M.~Asadi-Aghbolaghi, A.~Clapés, M.~Bellantonio, H.~J. Escalante,
  V.~Ponce-López, X.~Baró, I.~Guyon, S.~Kasaei, and S.~Escalera,
\newblock ``A survey on deep learning based approaches for action and gesture
  recognition in image sequences,''
\newblock in {\em 2017 12th IEEE International Conference on Automatic Face
  Gesture Recognition (FG 2017)}, May 2017, pp. 476--483.

\bibitem{integral-imaging}
V.~J. Traver, P.~Latorre-Carmona, E.~Salvador-Balaguer, F.~Pla, and B.~Javidi,
\newblock ``Three-dimensional integral imaging for gesture recognition under
  occlusions,''
\newblock {\em IEEE Signal Processing Letters}, vol. 24, no. 2, pp. 171--175,
  Feb 2017.

\bibitem{compressive-sensing}
H~Zhuang, M~Yang, Z~Cui, and Q~Zheng,
\newblock ``A method for static hand gesture recognition based on non-negative
  matrix factorization and compressive sensing,''
\newblock vol. 44, pp. 52--59, 01 2017.

\bibitem{temporal}
M.~Madadi, S.~Escalera, A.~Carruesco, C.~Andujar, X.~Baró, and J.~Gonzàlez,
\newblock ``Occlusion aware hand pose recovery from sequences of depth
  images,''
\newblock in {\em 2017 12th IEEE International Conference on Automatic Face
  Gesture Recognition (FG 2017)}, May 2017, pp. 230--237.

\bibitem{spatiotemporal}
Joshua Owoyemi and Koichi Hashimoto,
\newblock ``Spatiotemporal learning of dynamic gestures from 3d point cloud
  data,''
\newblock {\em CoRR}, vol. abs/1804.08859, 2018.

\bibitem{sensor-noise}
Raffaele Gravina, Parastoo Alinia, Hassan Ghasemzadeh, and Giancarlo Fortino,
\newblock ``Multi-sensor fusion in body sensor networks: State-of-the-art and
  research challenges,''
\newblock {\em Information Fusion}, vol. 35, pp. 68 -- 80, 2017.

\bibitem{multisensor}
Samarjit Das Inkyu~Moon Tabassum~Nasrin, Faliu~Yi,
\newblock ``Partially occluded object reconstruction using multiple kinect
  sensors,'' 2014.

\bibitem{leap}
G.~Marin, F.~Dominio, and P.~Zanuttigh,
\newblock ``Hand gesture recognition with leap motion and kinect devices,''
\newblock in {\em 2014 IEEE International Conference on Image Processing
  (ICIP)}, Oct 2014, pp. 1565--1569.

\bibitem{multisensor2}
K.~Liu, C.~Chen, R.~Jafari, and N.~Kehtarnavaz,
\newblock ``Fusion of inertial and depth sensor data for robust hand gesture
  recognition,''
\newblock {\em IEEE Sensors Journal}, vol. 14, no. 6, pp. 1898--1903, June
  2014.

\bibitem{pointnet}
Charles~Ruizhongtai Qi, Hao Su, Kaichun Mo, and Leonidas~J. Guibas,
\newblock ``Pointnet: Deep learning on point sets for 3d classification and
  segmentation,''
\newblock {\em CoRR}, vol. abs/1612.00593, 2016.

\bibitem{hand-pointnet}
Liuhao Ge, Yujun Cai, Junwu Weng, and Junsong Yuan,
\newblock ``Hand pointnet: 3d hand pose estimation using point sets,''
\newblock in {\em CVPR}, 2018.

\bibitem{noaugmentation1}
Liuhao Ge, Zhou Ren, and Junsong Yuan,
\newblock ``Point-to-point regression pointnet for 3d hand pose estimation,''
\newblock in {\em ECCV}, 2018.

\bibitem{pointnet++}
Charles~Ruizhongtai Qi, Li~Yi, Hao Su, and Leonidas~J. Guibas,
\newblock ``Pointnet++: Deep hierarchical feature learning on point sets in a
  metric space,''
\newblock {\em CoRR}, vol. abs/1706.02413, 2017.

\bibitem{frustum}
Charles~R Qi, Wei Liu, Chenxia Wu, Hao Su, and Leonidas~J Guibas,
\newblock ``Frustum pointnets for 3d object detection from rgb-d data,''
\newblock {\em arXiv preprint arXiv:1711.08488}, 2017.

\bibitem{integrating}
Jiaming Li, Yulan Guo, Yanxin Ma, Min Lu, and Jun Zhang,
\newblock ``Integrating color and depth cues for static hand gesture
  recognition,''
\newblock 11 2017, pp. 295--306.

\bibitem{open3d}
Qian{-}Yi Zhou, Jaesik Park, and Vladlen Koltun,
\newblock ``Open3d: {A} modern library for 3d data processing,''
\newblock {\em CoRR}, vol. abs/1801.09847, 2018.

\bibitem{Park2017ColoredPC}
Jaesik Park, Qian-Yi Zhou, and Vladlen Koltun,
\newblock ``Colored point cloud registration revisited,''
\newblock {\em 2017 IEEE International Conference on Computer Vision (ICCV)},
  pp. 143--152, 2017.

\bibitem{CHEN1992145}
Yang Chen and Gérard Medioni,
\newblock ``Object modelling by registration of multiple range images,''
\newblock {\em Image and Vision Computing}, vol. 10, no. 3, pp. 145 -- 155,
  1992,
\newblock Range Image Understanding.

\bibitem{arm-removal}
Bingyuan Xu, Zhiheng Zhou, Xi~Chen, Yi~Yang, and Zhiwei Yang,
\newblock ``Arm removal for static hand gesture recognition,''
\newblock {\em Journal of Intelligent \& Fuzzy Systems}, pp. 1--12, 11 2018.

\bibitem{augmentation1}
D.~Maturana and S.~Scherer,
\newblock ``Voxnet: A 3d convolutional neural network for real-time object
  recognition,''
\newblock in {\em 2015 IEEE/RSJ International Conference on Intelligent Robots
  and Systems (IROS)}, Sep. 2015, pp. 922--928.

\bibitem{augmentation2}
Lyne P.~Tchapmi, Chris Choy, Iro Armeni, JunYoung Gwak, and Silvio Savarese,
\newblock ``Segcloud: Semantic segmentation of 3d point clouds,''
\newblock 10 2017.

\bibitem{stn3d-git}
Shubham Tulsiani,
\newblock ``stn3d,'' https://github.com/shubhtuls/stn3d, 2018.

\bibitem{stn3d}
Max Jaderberg, Karen Simonyan, Andrew Zisserman, and Koray Kavukcuoglu,
\newblock ``Spatial transformer networks,''
\newblock {\em CoRR}, vol. abs/1506.02025, 2015.

\bibitem{pytorch-git}
Fei Xia,
\newblock ``pointnet.pytorch,'' https://github.com/fxia22/pointnet.pytorch,
  2019.

\bibitem{my-git}
Ilya Chugunov,
\newblock ``Duodepth,'' https://github.com/Ilya-Muromets/DuoDepth, 2018.

\bibitem{projection}
Chaoyu Liang, Yonghong Song, and Yuanlin Zhang,
\newblock ``Hand gesture recognition using view projection from point cloud,''
\newblock 09 2016, pp. 4413--4417.

\bibitem{albanian}
Eriglen Gani and Alda Kika,
\newblock ``Albanian sign language (albsl) number recognition from both
  hand’s gestures acquired by kinect sensors,''
\newblock {\em International Journal of Advanced Computer Science and
  Applications}, vol. 7, 08 2016.

\bibitem{SELU}
G{\"{u}}nter Klambauer, Thomas Unterthiner, Andreas Mayr, and Sepp Hochreiter,
\newblock ``Self-normalizing neural networks,''
\newblock {\em CoRR}, vol. abs/1706.02515, 2017.

\bibitem{attention}
Cherdsak Kingkan, Joshua Owoyemi, and Koichi Hashimoto,
\newblock ``Point attention network for gesture recognition using point cloud
  data,''
\newblock in {\em BMVC}, 2018.

\end{thebibliography}
}

\vspace{1em}
\footnotesize{$\copyright$ 2019 IEEE. Personal use of this material is permitted. Permission from IEEE must be obtained for all other uses, in any current or future media, including reprinting/republishing this material for advertising or promotional purposes, creating new collective works, for resale or redistribution to servers or lists, or reuse of any copyrighted component of this work in other works.}
\end{document}